\documentclass[9pt,twocolumn]{extarticle}

\usepackage[utf8]{inputenc}
\usepackage{graphicx}
\usepackage{amsmath}
\usepackage{amssymb}
\usepackage{booktabs}
\usepackage{multirow}
\usepackage{hyperref}
\usepackage{xcolor}
\usepackage[margin=1in]{geometry}
\usepackage{caption}
\usepackage{subcaption}
\usepackage{float}

\title{FOTBCD: A Large-Scale Building Change Detection Benchmark from French Orthophotos and Topographic Data\\
}

\author{
    Abdelrrahman Moubane \\
    Retgen AI \\
    \texttt{abdel@retgen.ai}
}

\date{}

\begin{document}

\maketitle

\begin{abstract}
We introduce FOTBCD, a large-scale building change detection dataset derived from authoritative French orthophotos and topographic building data provided by IGN France. Unlike existing benchmarks that are geographically constrained to single cities or limited regions, FOTBCD spans 28 departments across mainland France, with 25 used for training and three geographically disjoint departments held out for evaluation. The dataset covers diverse urban, suburban, and rural environments at 0.2\,m/pixel resolution.

We publicly release \textbf{FOTBCD-Binary}, a dataset comprising approximately 28,000 before/after image pairs with pixel-wise binary building change masks, each associated with patch-level spatial metadata. The dataset is designed for large-scale benchmarking and evaluation under geographic domain shift, with validation and test samples drawn from held-out departments and manually verified to ensure label quality.

In addition, we publicly release \textbf{FOTBCD-Instances}, a publicly available instance-level annotated subset comprising several thousand image pairs, which illustrates the complete annotation schema used in the full instance-level version of FOTBCD.

Using a fixed reference baseline, we benchmark FOTBCD-Binary against LEVIR-CD+ and WHU-CD, providing strong empirical evidence that geographic diversity at the dataset level is associated with improved cross-domain generalization in building change detection.
\end{abstract}

\section{Introduction}

Building change detection from remote sensing imagery is essential for urban planning, cadastral updates, disaster response, and environmental monitoring. Deep learning has driven significant progress, but model development and evaluation remain constrained by the limitations of existing datasets.

Current change detection benchmarks suffer from two critical limitations:

\textbf{Geographic Homogeneity.} LEVIR-CD+~\cite{levir} covers 20 regions in Texas, USA, while WHU-CD~\cite{whu} consists of a single image pair from Christchurch, New Zealand. Models trained on such geographically constrained data learn region-specific features---particular building styles, urban layouts, vegetation patterns, and imaging conditions---that fail to generalize across geographic domains.

\textbf{Limited Scale and Diversity.} Existing datasets contain hundreds to thousands of image pairs, insufficient to capture the full variability of real-world change detection scenarios, including dense urban cores, suburban sprawl, rural villages, industrial zones, and varying terrain.

We address these limitations with FOTBCD, a dataset family derived from France's authoritative geographic databases. In this work, we focus on the public release of \textbf{FOTBCD-Binary}, a large-scale benchmark for binary building change detection:

\begin{itemize}
    \item \textbf{National Coverage}: 28 departments across mainland France (25 for training and 3 held out for evaluation), spanning diverse geographic and urban contexts from Mediterranean coastal regions to rural Atlantic areas.
    \item \textbf{High Resolution}: 0.2\,m/pixel aerial orthophotos from IGN's BD ORTHO database.
    \item \textbf{Binary Change Annotations}: Pixel-wise CHANGE / NO-CHANGE masks suitable for large-scale benchmarking.
    \item \textbf{Spatially Referenced}: All patches are associated with patch-level spatial metadata, including projected-coordinate bounding boxes, enabling GIS integration.
    \item \textbf{Scale}: Approximately 28,000 image pairs yielding over 100,000 training 256x256 patches.
\end{itemize}

In addition, we release \textbf{FOTBCD-Instances (Research Dataset)}, a smaller subset with instance-level polygon annotations distinguishing NEW, DEMOLISHED, and UNCHANGED buildings, illustrating the full annotation schema and enabling instance-level experimentation.

Our experiments demonstrate that geographic diversity is a key factor for cross-domain generalization. Models trained on geographically limited datasets (LEVIR-CD+ and WHU-CD) exhibit substantial performance degradation when evaluated on FOTBCD-Binary. In contrast, a model trained on FOTBCD-Binary transfers effectively to WHU-CD and comparably to LEVIR-CD+ in relative terms, despite differences in imagery source and regional characteristics.

\section{Related Work}

\subsection{Change Detection Datasets}

Table~\ref{tab:dataset_overview} summarizes widely used building change detection datasets.

\textbf{LEVIR-CD}~\cite{levir} contains 637 pairs of 1024$\times$1024 Google Earth images at 0.5\,m resolution, covering approximately 20 regions in Texas, USA. Changes are annotated as binary masks focusing on building construction. The dataset is widely used but geographically homogeneous.

\textbf{WHU-CD}~\cite{whu} provides a single large pair of aerial images (32,507$\times$15,354 pixels at 0.2\,m resolution) from Christchurch, New Zealand, captured before and after the 2011 earthquake. While high-resolution, it represents only one city and a single change event.

\textbf{LEVIR-CD+} extends LEVIR-CD with additional image pairs while maintaining the same geographic scope and annotation format.

In contrast, \textbf{FOTBCD-Binary} provides large-scale geographic diversity at the national level, enabling evaluation across a wide range of urban, suburban, and rural environments. In addition, we release \textbf{FOTBCD-Instances}, a smaller subset with instance-level polygon annotations distinguishing NEW, DEMOLISHED, and UNCHANGED buildings, illustrating the full annotation schema.
\begin{table*}[t]
\centering
\caption{Comparison of building change detection datasets. FOTBCD-Binary provides national-scale geographic diversity, while FOTBCD-Instances illustrates instance-level annotations.}
\label{tab:dataset_overview}
\resizebox{\textwidth}{!}{%
\begin{tabular}{@{}lcccccccc@{}}
\toprule
\textbf{Dataset} & \textbf{Region} & \textbf{Coverage} & \textbf{Res.} & \textbf{Image pairs} & \textbf{Patch size} & \textbf{Ann.} & \textbf{Class} \\
\midrule
LEVIR-CD+ & Texas, USA & 20 areas & 0.5 m & 985 & 1024$\times$1024 & Mask & 2 \\
WHU-CD & Christchurch, NZ & 1 city & 0.2 m & 1 & 32,507×15,354 & Mask & 2 \\
\midrule
\textbf{FOTBCD-Binary} & \textbf{France} & \textbf{28 depts.} & \textbf{0.2 m} & \textbf{27871} & \textbf{512x512} & \textbf{Mask} & \textbf{2} \\
\textbf{FOTBCD-Instances} & \textbf{France} & \textbf{6 depts.} & \textbf{0.2 m} & \textbf{4000} & \textbf{512x512} & \textbf{Polygon} & \textbf{3} \\
\bottomrule
\end{tabular}}
\end{table*}

\subsection{Geographic Generalization in Remote Sensing}

Domain shift between training and deployment regions is a well-known challenge in remote sensing~\cite{domain_shift}. Variations in building architecture, urban density, land cover, terrain, and imaging conditions often cause models to underperform when applied to geographically distinct areas. Prior work has primarily addressed this problem through domain adaptation and normalization techniques. In contrast, we argue that increasing geographic diversity at the dataset level is a more fundamental and scalable approach to improving cross-region generalization.

\section{The FOTBCD Dataset}

\subsection{Data Sources}

FOTBCD is derived from two authoritative geospatial databases maintained by the Institut National de l'Information G\'{e}ographique et Foresti\`{e}re (IGN), France's national mapping agency.

\textbf{BD ORTHO} \cite{IGN_BD_ORTHO_2026} provides seamless aerial orthophoto coverage of mainland France at a ground sampling distance of 0.2 \,m. The dataset consists of RGB imagery acquired through regular national aerial survey campaigns, with update cycles ranging from 3–4 years depending on the region. The imagery is radiometrically corrected and orthorectified to ensure geometric consistency across large areas.

\textbf{BD TOPO} \cite{IGN_BD_TOPO_2026} is France’s authoritative topographic vector database, providing a structured 3D description of territorial elements and infrastructures with metric precision. The dataset covers the entire national territory and is exploitable at scales ranging from 1:2{,}000 to 1:50{,}000. It includes building footprints and associated semantic attributes, along with hydrographic, transport, administrative, land cover, and regulatory features.

Temporal differences between successive BD TOPO snapshots are analyzed and aligned with the corresponding BD ORTHO imagery to identify building-level changes, including \textit{new construction}, \textit{demolition}, and \textit{unchanged} structures. This change inference process forms the basis for both the large-scale binary change annotations released in \textbf{FOTBCD-Binary} and the instance-level polygon annotations provided in \textbf{FOTBCD-Instances}.

\subsection{Geographic Coverage}

\begin{figure}[H]
\centering
\includegraphics[width=\columnwidth]{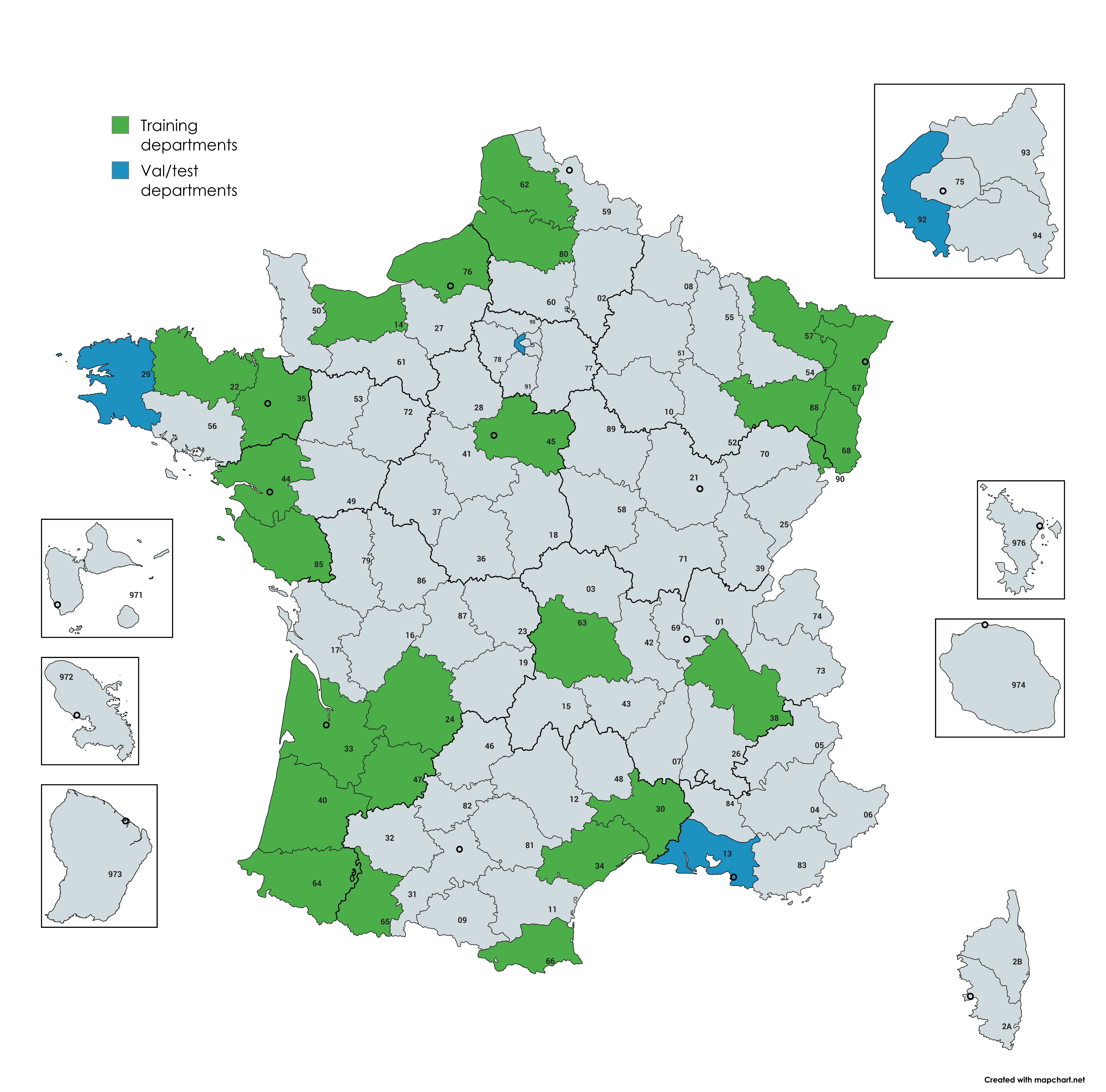}
\caption{Geographic coverage of FOTBCD-Binary. Training departments (25) are shown in green; held-out departments (3) used for validation and testing are shown in blue.}
\label{fig:coverage}
\end{figure}

FOTBCD-Binary spans \textbf{28 departments across mainland France}, with \textbf{25 departments used for training} and \textbf{3 geographically disjoint departments held out for validation and testing} (Figure~\ref{fig:coverage}). The split is performed at the department level to ensure zero geographic overlap between training and evaluation data, enabling an unbiased assessment of cross-region generalization.

The selected departments were chosen to maximize geographic, environmental, and urban diversity. Together, they cover a wide range of French landscapes, including:

\begin{itemize}
    \item \textbf{Dense urban and peri-urban regions}, encompassing large metropolitan areas and their surroundings
    \item \textbf{Coastal departments} along both the Atlantic Ocean and the Mediterranean Sea
    \item \textbf{Rural and agricultural areas}, including plains and low-density regions
    \item \textbf{Mountainous and foothill regions}, covering parts of the Alps and the Pyrenees
    \item \textbf{Industrial and mixed-use regions} in northern and eastern France
\end{itemize}

This geographic diversity results in substantial variability in:
\begin{itemize}
    \item Building typologies (dense apartment blocks, detached houses, industrial facilities, agricultural buildings)
    \item Urban density and spatial layout
    \item Surrounding land cover, including vegetation, water bodies, and terrain
    \item Imaging conditions driven by regional climate and acquisition campaigns
\end{itemize}

By spanning a broad cross-section of mainland France and enforcing a strict geographic hold-out protocol, FOTBCD-Binary encourages models to learn general change detection cues rather than region-specific appearance patterns.

\begin{table}[t]
\centering
\caption{FOTBCD-Binary dataset statistics.}
\label{tab:dataset_bianry_stats}
\resizebox{\columnwidth}{!}{%
\begin{tabular}{@{}lcccc@{}}
\toprule
\textbf{Department} & \textbf{Area (km²)} & \textbf{Patch Pairs} & \textbf{T1} & \textbf{T2} \\
\midrule
13 & 7.1 & 680 & 2014 & 2023 \\
14 & 10.2 & 968 & 2016 & 2023 \\
22 & 11.1 & 1054 & 2015 & 2021 \\
24 & 14.6 & 1393 & 2015 & 2024 \\
29 & 7.5 & 713 & 2015 & 2024 \\
30 & 7.6 & 721 & 2015 & 2024 \\
33 & 16.3 & 1556 & 2015 & 2024 \\
34 & 8.3 & 789 & 2015 & 2024 \\
35 & 13.7 & 1303 & 2014 & 2023 \\
38 & 11.6 & 1106 & 2015 & 2024 \\
40 & 13.0 & 1240 & 2015 & 2024 \\
44 & 15.2 & 1449 & 2016 & 2022 \\
45 & 12.2 & 1160 & 2016 & 2023 \\
47 & 13.0 & 1244 & 2015 & 2024 \\
57 & 12.1 & 1151 & 2015 & 2022 \\
62 & 11.3 & 1073 & 2015 & 2024 \\
63 & 14.0 & 1338 & 2016 & 2022 \\
64 & 13.4 & 1278 & 2015 & 2024 \\
65 & 4.0 & 385 & 2016 & 2022 \\
66 & 5.1 & 489 & 2015 & 2024 \\
67 & 8.2 & 781 & 2015 & 2024 \\
68 & 6.6 & 633 & 2015 & 2024 \\
76 & 8.7 & 834 & 2015 & 2022 \\
80 & 9.1 & 864 & 2017 & 2024 \\
85 & 21.1 & 2014 & 2016 & 2022 \\
88 & 7.5 & 717 & 2014 & 2023 \\
90 & 3.9 & 368 & 2017 & 2023 \\
92 & 6.0 & 570 & 2014 & 2024 \\
\midrule
\textbf{Total} & \textbf{292.2} & \textbf{27871} & - & - \\
\bottomrule
\end{tabular}}
\end{table}

\begin{table}[t]
\centering
\caption{FOTBCD-Instances dataset statistics.}
\label{tab:dataset_instances_stats}
\resizebox{\columnwidth}{!}{%
\begin{tabular}{@{}lcccc@{}}
\toprule
\textbf{Department} & \textbf{Area (km²)} & \textbf{Patch Pairs} & \textbf{T1} & \textbf{T2} \\
\midrule
13 & 3.5 & 330 & 2014 & 2023 \\
29 & 3.9 & 376 & 2015 & 2024 \\
47 & 12.0 & 1142 & 2015 & 2024 \\
63 & 12.7 & 1208 & 2016 & 2022 \\
88 & 6.8 & 650 & 2014 & 2023 \\
92 & 3.1 & 294 & 2014 & 2024 \\
\midrule
\textbf{Total} & \textbf{41.9} & \textbf{4000} & - & - \\
\bottomrule
\end{tabular}}
\end{table}
\subsection{Annotation Schema}

FOTBCD is released in two complementary variants with distinct annotation granularities.

\textbf{FOTBCD-Binary}, the primary public benchmark used for all experiments in this work, provides pixel-wise binary change annotations indicating the presence or absence of building change between two temporal images. These binary masks are derived from building-level change inference and are designed for large-scale benchmarking and cross-domain generalization studies.

In addition, we release \textbf{FOTBCD-Instances}, a smaller subset with instance-level polygon annotations in COCO format. This subset illustrates the full annotation schema and supports instance-level analysis and experimentation. Each building instance is assigned one of three semantic classes:
\begin{enumerate}
    \item \textbf{UNCHANGED} (category\_id=1): Buildings present in both temporal images.
    \item \textbf{DEMOLISHED} (category\_id=2): Buildings present in the ``before'' image but absent in the ``after'' image.
    \item \textbf{NEW} (category\_id=3): Buildings absent in the ``before'' image but present in the ``after'' image.
\end{enumerate}

For binary change detection, the \textbf{NEW} and \textbf{DEMOLISHED} classes are merged into a single CHANGE category, while UNCHANGED buildings are ignored.

The instance-level polygon representation enables precise boundary evaluation beyond pixel-wise metrics, instance-level performance analysis, and future research on multi-class and instance-aware change detection.
All image patches are provided with their original metadata georeferences in the Lambert-93 (EPSG:2154) coordinate system.

\subsection{Data Splits and Quality Control}

FOTBCD-Binary employs a strict geographic split strategy to ensure unbiased evaluation under cross-region domain shift.

\textbf{Training Set.} Training data are derived automatically from temporal differences in BD TOPO, aligned with corresponding BD ORTHO imagery. Building-level change signals are inferred using a combination of rule-based filtering, domain-specific heuristics, and learned components, and subsequently converted into binary change masks. As with most large-scale automatically generated datasets, the training set may contain limited label noise. However, our experimental results indicate that this noise does not materially affect model performance.

\textbf{Validation and Test Sets.} Validation and test data are sourced from \textbf{three departments that are completely excluded from the training set}, ensuring zero geographic overlap. All validation and test samples are \textbf{manually selected and verified by human annotators}, providing a reliable evaluation benchmark despite potential noise in the training data.

The quality control pipeline includes:
\begin{enumerate}
    \item \textbf{Temporal Alignment}: Ensuring consistency between BD ORTHO acquisition dates and BD TOPO snapshot dates.
    \item \textbf{Topological and Semantic Validation}: Detailed analysis of BD TOPO building geometry and metadata, followed by the application of domain-specific rules to remove implausible or inconsistent changes.
    \item \textbf{AI-Based Filtering}: Automated detection of residual inconsistencies and rare edge cases.
    \item \textbf{Human Verification}: Manual inspection of all validation and test samples to eliminate remaining label errors.
\end{enumerate}
\subsection{Qualitative Examples}
\label{sec:qualitative}

\begin{figure*}[t]
    \centering
    \includegraphics[width=\textwidth]{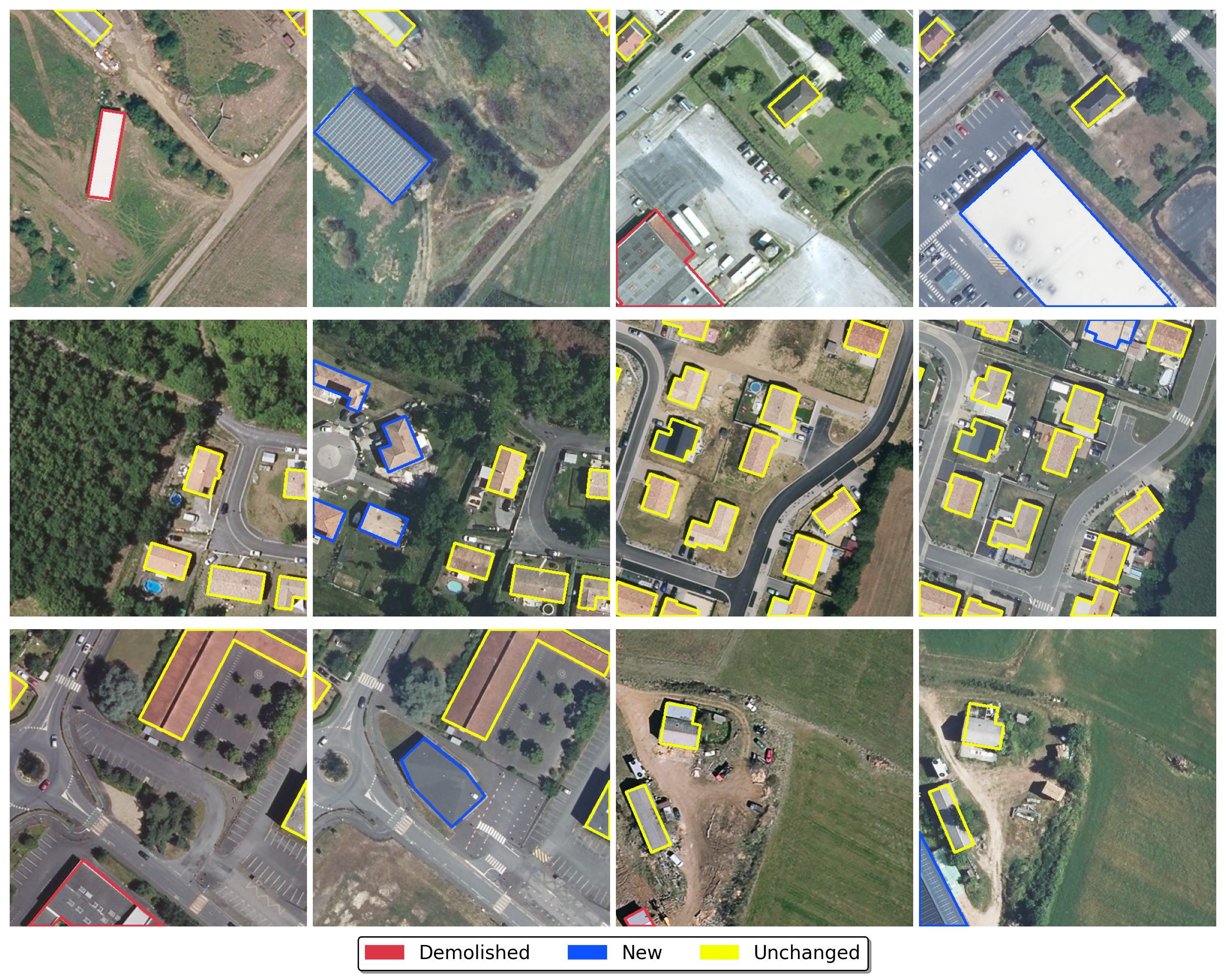}
    \caption{Qualitative visualization grid from FOTBCD.}
    \label{fig:qualitative_examples}
\end{figure*}

Figure~\ref{fig:qualitative_examples} presents a grid of qualitative
visualizations from \textbf{FOTBCD}. The figure is arranged
as a grid with two before/after image pairs per row. For each pair, building
polygons are drawn to illustrate the visual interpretation of building changes
between the two temporal images.

Polygon overlays follow a visualization-specific convention: \textit{DEMOLISHED}
buildings are drawn only on the ``before'' image, \textit{NEW} buildings are
drawn only on the ``after'' image, and \textit{UNCHANGED} buildings are drawn on
both images. This convention is used solely for visual clarity and does not
reflect the underlying annotation storage format.

\section{HybridSiam-CD: Reference Baseline}

To benchmark FOTBCD-Binary, we use HybridSiam-CD, a reference change detection model that combines pretrained Vision Transformer features with CNN-based boundary refinement. The model is not proposed as a new architecture, but is used solely to provide a consistent and reasonably strong baseline for cross-dataset evaluation.

\subsection{Architecture}

HybridSiam-CD follows a siamese design with separate semantic and spatial processing branches:

\begin{itemize}
    \item \textbf{Semantic Encoder}: A frozen DINOv3-sat493M~\cite{dinov3} Vision Transformer pretrained on satellite imagery extracts high-level semantic features from both temporal images. Absolute feature differences are used to encode change information.
    \item \textbf{Spatial Branch}: A siamese ResNet34 pretrained on ImageNet provides multi-scale spatial and boundary cues.
    \item \textbf{Fusion Decoder}: A lightweight decoder fuses semantic and spatial features using skip connections and upsampling to produce dense change maps.
\end{itemize}

The Vision Transformer encoder is kept frozen during training, resulting in a limited number of trainable parameters. This design prioritizes stability and reproducibility over architectural optimization.

\subsection{Training Protocol}

To ensure a consistent evaluation protocol across datasets of different sizes, all models are trained using the same fixed configuration:

\begin{itemize}
    \item \textbf{Training Schedule}: 50{,}000 optimization steps
    \item \textbf{Batch Size}: 128 patches of size $256\times256$
    \item \textbf{Optimizer}: AdamW with a learning rate of $4\times10^{-4}$
    \item \textbf{Learning Rate Schedule}: Cosine annealing with 2{,}000 warmup steps
    \item \textbf{Loss Function}: Combination of Lovasz hinge loss and boundary-aware BCE
    \item \textbf{Data Augmentation}: Random flips, rotations, and brightness/contrast adjustments
\end{itemize}

\section{Experiments}

\subsection{Cross-Domain Generalization}

The primary experiment evaluates cross-dataset generalization under geographic domain shift. Table~\ref{tab:cross_domain} reports Intersection-over-Union (IoU) scores when training on one dataset and evaluating on another.

\begin{table}[t]
\centering
\caption{Cross-domain generalization (IoU). Rows indicate the training dataset and columns the evaluation dataset's test split. Diagonal entries correspond to in-domain performance.}
\label{tab:cross_domain}
\small
\resizebox{\columnwidth}{!}{%
\begin{tabular}{@{}lccc@{}}
\toprule
\textbf{Train $\downarrow$ / Test $\rightarrow$} & \textbf{FOTBCD-Binary} & \textbf{LEVIR-CD+} & \textbf{WHU-CD} \\
\midrule
FOTBCD-Binary    & \textit{0.8180} & 0.2984 & 0.6974 \\
LEVIR-CD+ & 0.3003 & \textit{0.7365} & 0.5432 \\
WHU-CD    & 0.3417 & 0.2127 & \textit{0.8939} \\

\bottomrule
\end{tabular}}
\end{table}

Key observations:

\textbf{Generalization to FOTBCD-Binary.} Models trained on geographically constrained datasets exhibit substantial performance degradation when evaluated on FOTBCD-Binary. LEVIR-CD+ and WHU-CD achieve IoU scores of 0.3003 and 0.3417, respectively, indicating limited transfer to a more geographically diverse, national-scale dataset.

\textbf{Generalization from FOTBCD-Binary.} A model trained on FOTBCD-Binary generalizes well to WHU-CD, achieving an IoU of 0.6974, and performs comparably to the inverse transfer when evaluated on LEVIR-CD+ (0.2984). While cross-dataset performance remains below in-domain results, these outcomes are consistently more balanced than those observed when transferring from geographically constrained datasets to FOTBCD-Binary.

\textbf{Asymmetric domain shift.} Cross-evaluation between LEVIR-CD+ and WHU-CD also results in large performance drops relative to in-domain training, confirming that geographic domain shift is a general challenge in building change detection and not specific to any single dataset.

\newpage
\section{Discussion}

\subsection{Why Geographic Diversity Matters}

Our experiments indicate that geographic diversity at the dataset level is a key factor for improving robustness to cross-region domain shift in building change detection. This observation is consistent with findings in other machine learning domains, where diversity in training data often has a greater impact on out-of-distribution performance than raw dataset size.

The 25 training departments included in FOTBCD-Binary span a wide range of geographic and environmental conditions, including:
\begin{itemize}
    \item Multiple climate regimes (oceanic, continental, Mediterranean, mountain, and semi-arid)
    \item Diverse architectural styles and settlement patterns
    \item Terrain variations ranging from lowland plains to mountainous regions
\end{itemize}

Exposure to this diversity encourages models to learn more general change-related cues, rather than overfitting to region-specific building appearances, layouts, or imaging conditions. This effect is reflected in the cross-dataset evaluation results, where models trained on geographically constrained benchmarks struggle to generalize to FOTBCD-Binary, while training on a more diverse national-scale dataset leads to more balanced transfer behavior.

\subsection{Limitations}

Despite its scale and geographic diversity, FOTBCD has limitations:
\begin{itemize}
    \item \textbf{National scope}: The dataset is limited to mainland France. While it spans diverse regions and environments, evaluation across multiple countries or continents would require additional datasets.
    \item \textbf{Building-centric annotations}: The annotations focus exclusively on building changes; other change categories such as roads, vegetation, or land use are not labeled.
\end{itemize}

\section{Conclusion}

We introduced FOTBCD, a national-scale building change detection dataset derived from France’s authoritative geographic data, designed to support robust evaluation under geographic domain shift. The dataset spans 25 training departments with three geographically disjoint departments reserved for validation and testing, covering a wide range of urban, suburban, and rural environments at 0.2\,m spatial resolution.

Using a fixed reference baseline, our experiments show that models trained on geographically diverse, national-scale data exhibit improved robustness to cross-dataset evaluation compared to models trained on geographically constrained benchmarks. In particular, models trained on FOTBCD-Binary generalize more consistently across datasets, while models trained on LEVIR-CD+ or WHU-CD experience substantial performance degradation when evaluated on FOTBCD-Binary.

FOTBCD is intended as a foundation for future research on geographically robust change detection. By releasing both a large-scale binary benchmark and a smaller instance-level dataset, we aim to support a broad range of research directions, from semantic change detection to instance-level analysis.

\section*{Data Availability and Licensing}

FOTBCD is released under a clear separation between publicly available research datasets and a larger, non-public dataset intended for commercial use.

\textbf{FOTBCD-Binary (Research Dataset).}  
The primary dataset used for training and evaluation in this paper is publicly released for research use under the \textbf{CC BY-NC-SA 4.0} license (\url{https://creativecommons.org/licenses/by-nc-sa/4.0/deed.en}). It includes:
\begin{itemize}
    \item approximately 28{,}000 before/after image pairs,
    \item pixel-wise binary building change masks derived from authoritative vector data,
    \item human-verified validation and test splits.
\end{itemize}

\textbf{FOTBCD-Instances (Research Subset).}  
A publicly available instance-level research subset, released under the same \textbf{CC BY-NC-SA 4.0} license, comprising approximately 4{,}000 before/after image pairs with dedicated validation and test splits. The dataset is annotated with building polygons labeled as \textit{NEW}, \textit{DEMOLISHED}, or \textit{UNCHANGED}, and supports instance-level training and evaluation as a complement to the binary benchmark.

\medskip
\noindent
\textbf{Repository Access.}  
Both public research datasets (FOTBCD-Binary and FOTBCD-Instances) are distributed through a single public GitHub repository together with training code, configuration files, and pretrained model checkpoints:  
\url{https://github.com/abdelpy/FOTBCD-datasets}.

\medskip
\noindent
\textbf{FOTBCD-220k (Internal Research Data).}  
In addition to the public research datasets, a larger instance-level dataset comprising over 220{,}000 image pairs has been constructed for internal research and commercial applications. This dataset is not publicly released and is distributed separately under a commercial license. Access may be granted to academic institutions within the context of collaborative research projects.

\medskip
\noindent
\newpage
\section*{Acknowledgments}

FOTBCD is derived from the BD ORTHO and BD TOPO databases made freely available by the Institut National de l'Information G\'{e}ographique et Foresti\`{e}re (IGN) under the Licence Ouverte / Open Licence version~2.0\footnote{\url{https://alliance.numerique.gouv.fr/licence-ouverte-open-licence/}}.

\bibliographystyle{plain}
\bibliography{references}

\begin{thebibliography}{1}

\bibitem{levir}
Hao Chen and Zhenwei Shi.
\newblock A spatial-temporal attention-based method and a new dataset for
  remote sensing image change detection.
\newblock {\em Remote Sensing}, 12(10), 2020.

\bibitem{IGN_BD_ORTHO_2026}
{Institut National de l’Information Géographique et Forestière (IGN)}.
\newblock {BD ORTHO}.
\newblock \url{https://geoservices.ign.fr/bdortho}.
\newblock Accessed: 2026-01-27.

\bibitem{IGN_BD_TOPO_2026}
{Institut National de l’Information Géographique et Forestière (IGN)}.
\newblock {BD TOPO}.
\newblock \url{https://geoservices.ign.fr/bdtopo}.
\newblock Accessed: 2026-01-27.

\bibitem{whu}
Shunping Ji, Shiqing Wei, and Meng Lu.
\newblock Fully convolutional networks for multisource building extraction from
  an open aerial and satellite imagery data set.
\newblock {\em IEEE Transactions on Geoscience and Remote Sensing},
  57(1):574--586, 2019.

\bibitem{dinov3}
Oriane Siméoni, Huy~V. Vo, Maximilian Seitzer, Federico Baldassarre, Maxime
  Oquab, Cijo Jose, Vasil Khalidov, Marc Szafraniec, Seungeun Yi, Michaël
  Ramamonjisoa, Francisco Massa, Daniel Haziza, Luca Wehrstedt, Jianyuan Wang,
  Timothée Darcet, Théo Moutakanni, Leonel Sentana, Claire Roberts, Andrea
  Vedaldi, Jamie Tolan, John Brandt, Camille Couprie, Julien Mairal, Hervé
  Jégou, Patrick Labatut, and Piotr Bojanowski.
\newblock Dinov3, 2025.

\bibitem{domain_shift}
Mei Wang and Weihong Deng.
\newblock Deep visual domain adaptation: {A} survey.
\newblock {\em CoRR}, abs/1802.03601, 2018.

\end{thebibliography}
\end{document}